\documentclass[12pt,linesnumbered,ruled,vlined]{article}
\usepackage{mathptmx}
\usepackage[latin9]{inputenc}
\usepackage{color}
\usepackage{mathrsfs}
\usepackage{mathtools}
\usepackage{amsmath}
\usepackage{amstext}
\usepackage{amsthm}
\usepackage{amssymb}
\usepackage{graphicx}
\usepackage{appendix}
\usepackage[unicode=true,pdfusetitle,
 bookmarks=true,bookmarksnumbered=false,bookmarksopen=false,
 breaklinks=false,pdfborder={0 0 1},backref=false,colorlinks=true]
 {hyperref}
\hypersetup{
 pdfborderstyle={},pdfborderstyle={},pdfborderstyle={},urlcolor=blue,linkcolor=blue,citecolor=blue}

\makeatletter

\DeclareTextSymbolDefault{\textquotedbl}{T1}

\numberwithin{equation}{section}
\numberwithin{figure}{section}
\theoremstyle{remark}
\newtheorem*{acknowledgement*}{\protect\acknowledgementname}

\@ifundefined{date}{}{\date{}}

\date{\today}

\usepackage{amsfonts}
\usepackage{dsfont}

\numberwithin{equation}{section}

\allowdisplaybreaks[4]
\DeclareMathAlphabet{\mathcal}{OMS}{cmsy}{m}{n}

\makeatother

\providecommand{\acknowledgementname}{Acknowledgement}

\begin{document}
\title{Deep Generative Modeling with Backward Stochastic Differential Equations}
\author{Xingcheng Xu\thanks{Shanghai AI Laboratory. Email: xingcheng.xu18@gmail.com}\ \ \footnote{We will make our code publicly available on GitHub at  \href{https://github.com/xingchengxu/BSDE-Gen}{https://github.com/xingchengxu/BSDE-Gen} after the completion
of our research. Please send an email to request access if you're interested, in case the access is unavailable.}}
\maketitle

\begin{abstract}
{\normalsize{}
This paper proposes a novel deep generative model, called
\textbf{\textit{BSDE-Gen}}, which combines
the flexibility of backward stochastic differential equations (BSDEs)
with the power of deep neural networks for generating high-dimensional
complex target data, particularly in the field of image generation.
The incorporation of stochasticity and uncertainty in the generative
modeling process makes BSDE-Gen an effective and natural approach
for generating high-dimensional data. The paper provides a theoretical
framework for BSDE-Gen, describes its model architecture, presents
the maximum mean discrepancy (MMD) loss function used for training,
and reports experimental results.
}{\normalsize\par}
\end{abstract}

\vspace{5mm}
 \hspace{5mm}\textbf{Keywords: }Backward Stochastic Differential
Equation, FBSDE, BSDE-Gen, 

\hspace{5mm}BSDE-based Generative Models, High-dimensional Learning,
Image Generation

\section{Introduction\label{sec:Introduction}}

Generative models are statistical models that aim to learn the probability
distribution of a dataset in order to generate new data that is similar
to the original data. In recent years, deep generative diffusion models
have become increasingly popular due to their ability to generate
high-quality data that closely resembles real data. Notable examples
of diffusion models include GLIDE, DALL-E 2, Imagen, and Stable Diffusion,
among others.

In this paper, we propose a novel approach for developing deep generative
models that combines the flexibility of BSDEs with the power of deep
neural networks. Our model, called BSDE-Gen, leverages the strengths
of BSDEs in describing the evolution of a stochastic process backwards
in time, starting from a given terminal condition, and deep neural
networks' ability to model complex, high-dimensional data.

The BSDE-Gen model starts with a random initial input that follows
a high-dimensional standard normal distribution and progresses towards
a final value that represents the target data distribution. Once trained
with a loss function, the optimized parameters can be used to generate
new samples using a forward scheme. During training, the model learns
the probability distribution of a given dataset by minimizing a MMD
loss function derived from the BSDE-based dynamics. As a result, the
model can generate new data that closely resembles the original dataset.

Our method is based on rigorous mathematical principles and theory.
The implications of this work could be significant for various fields
that rely on generative models to simulate or generate new data, such
as computer vision, biology, and drug discovery. By incorporating
BSDEs into deep generative models, it could provide a new tool for
modeling complex systems with uncertain dynamics and incomplete information.

The paper is organized as follows: Section \ref{sec:Related-work}
provides a review of relevant literature on diffusion models and BSDEs.
In Section \ref{sec:Methodology}, we introduce a deep generative
model based on BSDEs and explains the model architecture and loss
function used. Section \ref{sec:Experiments} presents the results
of experiments conducted on the MNIST and FashionMNIST datasets to
demonstrate the effectiveness of the BSDE-based generative models.
In Section \ref{sec:Discussion}, we discuss the limitations of the
model, along with potential avenues for future research. Finally,
Section \ref{sec:Conclusion} provides the conclusion of the paper.

\section{Related work\label{sec:Related-work}}

In this section, we present a review of the relevant literature in
the areas of deep generative models and BSDEs, which are related to
our work. Our review covers diffusion models that employ deep neural
networks for generative modeling, as well as BSDEs, deep BSDE methods,
their applications, and related topics.

\subsection{Diffusion Models}

Diffusion-based generative models use (discrete/continuous) stochastic
differential equations (SDEs) to generate complex data and have gained
popularity in image generation for their ability to produce high-quality
and diverse images. 

There are two main diffusion processes in these models. The forward
diffusion process gradually adds Gaussian noise to a high-quality
image in several steps, producing a sequence of noisy images. While
the reverse diffusion process removes noise from a noise vector to
generate a high-quality image. Multiple diffusion-based generative
models have been proposed over the years, including Diffusion Probabilistic
Models (Sohl-Dickstein et al., 2015) \cite{SWM2015}, Noise-Conditioned
Score Network (NCSN; Song and Ermon, 2019) \cite{SE2019}, and Denoising
Diffusion Probabilistic Models (DDPM; Ho et al., 2020) \cite{HJA2020}.
During training, the models learn transformations to remove noise
while retaining the underlying structure of the image. 

When training generative models on datasets, it's common to generate
samples conditioned on class labels or descriptive text. Recently,
Dhariwal and Nichol (2021) \cite{DN2021} demonstrated that diffusion
models outperform GANs in this regard. Other notable works in this
area include the guided diffusion model (GLIDE; Nichol et al., 2022)
\cite{NDR2021}, as well as DALL-E 2 (Ramesh et al., 2022) \cite{RDN2022},
Imagen (Saharia et al., 2022) \cite{SCS2022}, and Latent Diffusion
Models/Stable Diffusion (Rombach et al., 2022) \cite{RBL2022}, etc. 

For a comprehensive survey of methods and applications on generative
diffusion models, we refer to survey papers such as \cite{CTG+2022,CHIS2022,YZS+2022}. 

Our work introduces a novel approach to generative modeling by utilizing
diffusion processes, which differs from traditional diffusion models.
Rather than relying on denoising, we employ deep neural networks (DNNs)
to solve BSDEs. While our framework does not currently include a guided
model, we believe that future research could explore the integration
of guided models to further enhance the performance and quality of
generated data. 

\subsection{BSDEs and Deep BSDE Methods}

Linear BSDEs were first proposed by Bismut in 1973 \cite{Bis1973}
as an adjoint equation for the stochastic optimal control problem.
In 1990, Pardoux and Peng \cite{PP1990} established the existence
and uniqueness of nonlinear BSDEs with Lipschitz condition, which
led to further research and applications in fields such as stochastic
optimal control and mathematical finance. When a BSDE is coupled with
a forward stochastic differential equation (SDE), it forms a forward-backward
stochastic differential equation (FBSDE), see e.g. \cite{HP1995,MPY1994},
which has also been extensively studied by researchers and applied
in various contexts. 

BSDEs have numerous applications in mathematical finance, including
asset pricing, portfolio optimization, and risk management. Specifically,
they are widely used for modeling and pricing derivatives, which are
financial instruments whose value is dependent on the value of an
underlying asset. The use of BSDEs allows for the formulation of a
dynamic model that describes the underlying asset and its associated
derivative, enabling the derivation of prices that are consistent
with the asset's behavior, see e.g. \cite{CE2002,EKPQ1997}. This
technique has been applied to a diverse range of asset classes such
as stocks, bonds, currencies, and commodities, making it a valuable
tool in the financial industry. 

The deep BSDE method is an innovative approach that combines classical
BSDE theory with deep neural networks to approximate unknown functions
in equations. This method has shown promising results in solving FBSDEs
and partial differential equations (PDEs) with high-dimensional state
spaces, see e.g. \cite{EHJ2017,HJE2018,HL2020,HHL2022,JPPZ2020,JL2021},
which were previously considered computationally intractable. The
approach has been applied to various problems in stochastic optimal
control and finance, such as the investment-consumption problem, option
pricing and portfolio optimization, and has demonstrated significant
improvements in accuracy and computational efficiency compared to
traditional methods. Some examples of relevant literature include
\cite{CW2021,JPPZ2022,WN2022,YXS2019,YGH2023}. The deep BSDE method
represents a new direction in the study of FBSDEs and PDEs, and it
has the potential to be applied in diverse fields such as physics,
engineering, and machine learning.

Our work builds upon deep BSDE methods, with a focus on generative
modeling, particularly for image generation. While deep BSDE methods
have previously been utilized for solving high dimensional FBSDEs
or PDEs, our approach differs as we utilize them for generative modeling.
In contrast to the work of Han, Jentzen and E (2018) \cite{HJE2018},
our framework utilizes a single neural network for the control process
$Z$, with shared parameters for different time steps. Additionally,
we employ a maximum mean discrepancy (MMD) loss instead of the mean
squared error (MSE) loss used in their work. Another distinguishing
feature of our model is that we use a random input to initiate the
forward process, rather than a deterministic point. This requires
us to use a deep neural network to generate the initial value $Y_{0}$,
rather than relying on fixed model parameters. These differences demonstrate
the unique approach of our framework in image generation and deep
BSDE methods.

\section{Methodology\label{sec:Methodology}}

BSDEs are stochastic differential equations that involve a terminal
condition and a backward evolution in time. They describe the evolution
of a stochastic process backwards in time, in contrast to forward
SDEs which describe the evolution of a process forward in time. BSDEs
are useful for modeling and analyzing complex systems with uncertain
dynamics and incomplete information.

In this section, we will introduce a deep generative model based on
BSDEs, the model architecture, and the MMD loss function we used.

\subsection{BSDE-based Generative Models}

Let $(\Omega,\mathcal{F},\mathbb{P})$ be a complete probability space,
$W_{t}:[0,T]\times\Omega\to\mathbb{R}^{d_{W}}$ be a standard $d_{W}$-dimensional
Brownian motion for some given constant $T>0$. Denote $\mathcal{F}_{t}=\sigma\{W_{s},\ 0\leq s\leq t\}$
as the natural filtration generated by the standard Brownian motions
for any $t\in[0,T]$. Consider the forward-backward stochastic differential
equation (FBSDE)
\begin{equation}
\begin{aligned}X_{t} & =\zeta+\int_{0}^{t}b(s,X_{s})ds+\int_{0}^{t}\sigma(s,X_{s})dW_{s},\\
Y_{t} & =\xi+\int_{t}^{T}f(s,X_{s},Y_{s},Z_{s})ds-\int_{t}^{T}Z_{s}dW_{s},
\end{aligned}
\end{equation}
where $t\in[0,T]$, $X_{t}\in\mathbb{R}^{d_{X}}$ is the forward process,
$Y_{t}\in\mathbb{R}^{d_{Y}}$ is the backward process, $Z_{t}\in\mathbb{R}^{d_{Y}\times d_{W}}$
is the control process. The drift function $b(t,x)$ and diffusion function $\sigma(t,x)$
define the dynamics of the forward process $X_{t}$, while the generator
function $f(t,x,y,z)$ specifies the relationship between the forward
process and the backward process $Y_{t}$. The mappings in above FBSDE
are as follows: 
\[
b:[0,T]\times\mathbb{R}^{d_{X}}\to\mathbb{R}^{d_{X}},
\]
\[
\sigma:[0,T]\times\mathbb{R}^{d_{X}}\to\mathbb{R}^{d_{X}\times d_{W}},
\]
\[
f:[0,T]\times\mathbb{R}^{d_{X}}\times\mathbb{R}^{d_{Y}}\times\mathbb{R}^{d_{Y}\times d_{W}}\to\mathbb{R}^{d_{Y}}.
\]
$X_{0}=\zeta$ is a given initial condition, and $Y_{T}=\xi$
is the terminal condition. $Y_{T}$ could be a function of $X_{T}$,
that is, $Y_{T}=\varphi(X_{T})$. The case is called \textit{Markovian
FBSDE}. For the \textit{non-Markovian FBSDE}, the terminal condition
$Y_{T}=\varphi(X_{0\leq t\leq T})$, that is, the terminal condition
is a functional of the whole path of $X_{t}$. We assume that the
initial input of the forward process $\zeta$ is subjected to a normal
distribution, for example, $\mathscr{N}(0,I_{d_{X}})$, and the terminal
value of the backward process $\xi$ is subjected to the target data
distribution. 

Solving a FBSDE involves finding the $\mathcal{F}_{t}$-adapted stochastic
process $(X_{t},Y_{t},Z_{t})$ for all $t\in[0,T]$ in a suitable
space that satisfies the equation above, given the functions $b,\sigma,f$,
Brownian motion $W_{t}$, the initial condition $\zeta$ and the terminal
condition $\xi$. The first results on the existence and uniqueness
of the solution to nonlinear BSDEs was given by Pardoux and Peng (1990)
\cite{PP1990}. Since then, many researchers studied the existence,
uniqueness and the applications of coupled or fully-coupled FBSDEs,
see e.g. \cite{HP1995,MPY1994}.

The FBSDEs can be connected to semilinear parabolic PDEs through the
Feynman-Kac formula under appropriate conditions (see e.g. \cite{PP1992,PT1999}).
The processes $Y_{t}$ and $Z_{t}$ depend on the time variable $t$
and the forward process $X_{t}$, rather than the entire path of $X$.
Specifically, $Y_{t}=u(t,X_{t})$ and $Z_{t}=\nabla u(t,X_{t})^{T}\sigma(t,X_{t})$,
where $u(t,x)$ satisfies the corresponding PDE. This property is
advantageous in designing deep neural networks for solving FBSDE models.

As an example, the forward process $X_{t}$ can be taken as a $d_{W}$-dimensional
Brownian motion that commences from a standard normal distribution,
i.e.
\begin{equation}
X_{t}=\zeta+W_{t},
\end{equation}
 or as a $d_{W}$-dimensional Ornstein-Uhlenbeck (OU) process starting
from the stationary distribution, i.e. 
\[
X_{t}=\zeta-\int_{0}^{t}\varLambda X_{s}ds+\Sigma W_{t}.
\]
The OU process has a unique solution under suitable conditions: 
\begin{equation}
X_{t}=e^{-\Lambda t}\zeta+\int_{0}^{t}e^{-\Lambda(t-s)}\Sigma dW_{s}.
\end{equation}
Suppose $\Lambda$ is a positive definite matrix. Then, $X$ has a
unique stationary distribution which is Gaussian with mean 0 and covariance
$\int_{0}^{\infty}e^{-\Lambda s}\Sigma\Sigma^{T}e^{-\Lambda^{T}s}ds$.
The initial value $\zeta$ can be randomly selected from the stationary
distribution of the forward process $X_{t}$.

The backward process $Y_{t}$ can be modeled, for example, as
\begin{equation}
Y_{t}=\xi+\int_{t}^{T}(AX_{s}+BY_{s}+\kappa|Z_{s}|)ds-\int_{t}^{T}Z_{s}dW_{s},
\end{equation}
in which the generator function is linear in $X_{t}$, $Y_{t}$, nonlinear
in $Z_{t}$ with the given matrix $A,B$ and $\kappa$, and $|z|:=(\sum_{j=1}^{d_{W}}|z_{ij}|)_{i=1,2,\cdots,d_{Y}}$. 

Generally, given a partition of the time interval $[0,T]$: $0=t_{0}<t_{1}<\cdots<t_{N}=T$,
using the naive \textit{Euler forward discrete scheme} for both the
forward and backward processes, we have 
\begin{equation}
\begin{aligned}X_{t_{n+1}} & \approx X_{t_{n}}+b(t_{n},X_{t_{n}})\Delta t_{n}+\sigma(t_{n},X_{t_{n}})\Delta W_{t_{n}}\\
Y_{t_{n+1}} & \approx Y_{t_{n}}-f(t_{n},X_{t_{n}},Y_{t_{n}},Z_{t_{n}})\Delta t_{n}+Z_{t_{n}}\Delta W_{t_{n}}
\end{aligned}
\label{eq:fbsde-discrete}
\end{equation}
where $\Delta t_{n}=t_{n+1}-t_{n}$ and $\Delta W_{t_{n}}=W_{t_{n+1}}-W_{t_{n}}$. 

\subsection{Model Architecture}

When the drift function $b(t,x)$ and diffusion function $\sigma(t,x)$
of the forward process and the generator function $f(t,x,y,z)$ are
given, by the Euler forward discrete scheme (\ref{eq:fbsde-discrete}),
what we should learn about the FBSDE is the initial value of the backward
process $Y_{0}$ and the control process $Z_{t}$. Since, under appropriate
mathematical conditions, $Y_{t}$ and $Z_{t}$ are functions of the
time variable $t$ and the forward process $X_{t}$, we can employ
two deep neural networks $\mathcal{N}^{\theta_{Y_{0}}}(X_{0})$ and
$\mathcal{N}^{\theta_{Z}}(t_{n},X_{t_{n}})$ to approximate the initial
value $Y_{0}$ and the control process $Z_{t}$, respectively. That
is,
\begin{equation}
Y_{0}\approx\mathcal{N}^{\theta_{Y_{0}}}(X_{0}),\ \ Z_{t_{n}}\approx\mathcal{N}^{\theta_{Z}}(t_{n},X_{t_{n}}),
\end{equation}
where the deep neural networks we used in our experiments in this
paper are as follows: 
\[
\mathcal{N}^{\theta}(x):=\phi\circ\mathcal{L}_{H}\circ\tilde{\sigma}_{H-1}\circ\mathcal{L}_{H-1}\circ\cdots\circ\tilde{\sigma}_{1}\circ\mathcal{L}_{1}(x),
\]
in which $H$ is the depth of the neural network, $\mathcal{L}_{h}(x_{h})=w_{h}x_{h}+\tilde{b}_{h}$
is the linear transformation, $\tilde{\sigma}_{h}$ is the nonlinear
activation function such as ReLU, Tanh, sigmoid, GELU, etc., and $\phi$
is the mapping function to the state space. As an additional regularization
technique, we utilize dropout in our neural networks. 

The overall architecture of the BSDE-Gen model is shown in Figure
\ref{model_architecture}, as it is suggested by the Euler forward
discrete scheme (\ref{eq:fbsde-discrete}). The BSDE-Gen model begins
by initializing a random input $X_{0}$ and a sequence of Brownian
motion $W_{t}$ at $N$ discrete time steps. The forward process $X_{t}$
is obtained using a discrete scheme of $X_{t}$. The initial value
$Y_{0}$ is obtained using a DNN $\mathcal{N}^{\theta_{Y_{0}}}(X_{0})$
that takes the input $X_{0}$ as input, and the control process $Z_{t}$
is obtained through another DNN $\mathcal{N}^{\theta_{Z}}(t,X_{t})$
as a function of the time variable $t$ and the forward process $X_{t}$.
By combining these elements, we can obtain $Y_{t}$ using the Euler
forward discrete scheme (\ref{eq:fbsde-discrete}) for $Y_{t}$ through
$N$ steps of diffusion. Finally, we obtain the generated image $Y_{T}$
at the final time $T$. The parameters $\theta=\{\theta_{Y_{0}},\theta_{Z}\}$
can be trained through the optimization of a loss function given in
the following subsection.


\subsection{The MMD Loss}

One possible candidate loss function could be the MMD. Assume there
is a feature map $\psi:\mathcal{X}\to\mathcal{H}$ from the original
space $\mathcal{X}$ to a Hilbert space $\mathcal{H}$, and the associated
kernel is a function $K:\mathcal{X}\times\mathcal{X}\to\mathbb{R}$
with the property that $\langle\psi(x),\psi(y)\rangle_{\mathcal{H}}=K(x,y)$
for all $x$ and $y$ in $\mathcal{X}$. The MMD computes the distance
between probability distributions as the distance between mean embeddings
of features via reproducing kernel Hilbert space (RKHS) $\mathcal{H}$.
Let $\text{\ensuremath{\mathbb{P}}}$ and $\mathbb{Q}$ be two probabilities
of random elements on the space $\mathcal{X}$, the MMD is defined
as 
\[
\textrm{MMD}^{2}(\mathbb{P},\mathbb{Q})=\|\mu_{\mathbb{P}}-\mu_{\mathbb{Q}}\|_{\mathcal{H}}^{2},
\]
where $\mu_{\mathbb{P}}=\mathbb{E}_{X\sim\mathbb{P}}[\psi(X)]$ and
$\mu_{\mathbb{Q}}=\mathbb{E}_{Y\sim\mathbb{Q}}[\psi(Y)]$ are the
mean embeddings of probabilities $\mathbb{P}$ and $\mathbb{Q}$ in
a RKHS $\mathcal{H}$, respectively. It is a kernel based statistical
test used to determine whether two given probability distributions
are the same. Under suitable conditions, $\textrm{MMD}^{2}(\mathbb{P},\mathbb{Q})=0$
if and only if $\mathbb{P}=\mathbb{Q}$. We refer to Gretton et al.
(2012) \cite{GBRS2012} for more details about the MMD, and Dziugaite
et al. (2015) \cite{DRG2015} for the MMD optimization.

The empirical estimate of the MMD for two samples $X=\{X_{1},\cdots,X_{m}\}$
and $Y=\{Y_{1},\cdots,Y_{n}\}$ is given as follows: 
\begin{equation}
\textrm{\ensuremath{\widehat{\textrm{MMD}}}}^{2}(X,Y)=\frac{1}{m(m-1)}\sum_{i=1}^{m}\sum_{j\neq i}^{m}K(X_{i},X_{j})+\frac{1}{n(n-1)}\sum_{i=1}^{n}\sum_{j\neq i}^{n}K(Y_{i},Y_{j})-\frac{2}{mn}\sum_{i=1}^{m}\sum_{j=1}^{n}K(X_{i},Y_{j}).
\end{equation}
The kernel $K(x,y)$ could be taken as a radial basis function (RBF)
kernel, multi-scaled kernel, etc. The MMD loss function for our BSDE-Gen
model is 
\begin{equation}
L(\theta)=\textrm{\ensuremath{\widehat{\textrm{MMD}}}}^{2}\left(Y_{T},\xi\right),
\end{equation}
where $Y_{T}$ is the final value computed by the Euler forward discrete
scheme (\ref{eq:fbsde-discrete}). In the training process, the loss
is calculated by a batch size $n_{batch}$ of the generated images
$Y_{T}$ and the target images $\xi$. 

The FBSDE dynamics begin with a $d_{X}$-dimensional initial input
$\zeta$ that follows a high dimensional standard normal distribution,
and evolve towards a final value $\xi$ that represents the target
data distribution. Once the optimized parameters $\hat{\theta}$ have
been obtained, new samples can be generated using the Euler forward
discrete scheme (\ref{eq:fbsde-discrete}).

\subsection{Training Strategies}

There are two training strategies for the BSDE-Gen models: decoder-only
style (shown in (a) of Figure \ref{training_strategy}) and encoder-decoder
style (shown in (b) of Figure \ref{training_strategy}). The decoder-only
style entails training the generative model to transform a noise vector
into a realistic image, similar to traditional Generative Adversarial
Networks (GANs; Goodfellow et al., 2014) \cite{Good2014}. In contrast,
the encoder-decoder style involves adding noise to the image, similar
to diffusion models (e.g. \cite{HJA2020,SWM2015,SE2019}). However,
in our paper, the purpose of this is not to simulate the reverse process
but rather to enhance the mapping between the BSDE-Gen model input and the target
image. We can achieve this by adding noise to the image, such as $\xi^{noisy}=\alpha\xi+(1-\alpha)\epsilon$,
where $\xi$ is the real image, $\epsilon\sim\mathscr{N}(0,I)$ and
$0<\alpha<1$. An encoder can also be used as an alternative to this. Under this case, the loss could be combined as $\textrm{Loss}=\textrm{MMD}+\beta\times\text{MSE}$.


\vspace*{\fill}
\begin{center}
\begin{figure}[tbph]
\begin{centering}
\includegraphics[width=15cm]{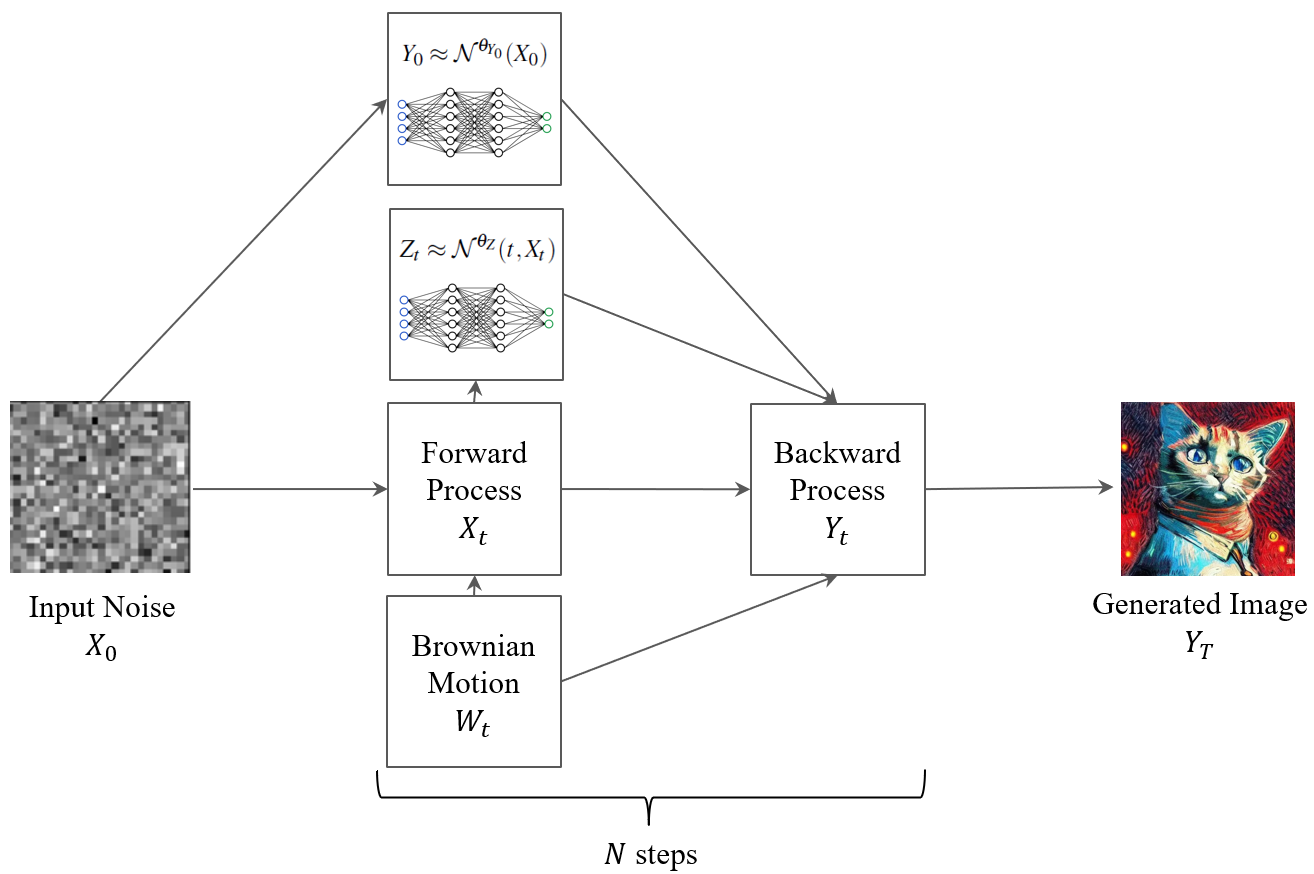}
\par\end{centering}
\caption{Model Architecture of BSDE-Gen}
\label{model_architecture}
\end{figure}
\begin{figure}[tbph]
\begin{centering}
\includegraphics[width=14cm]{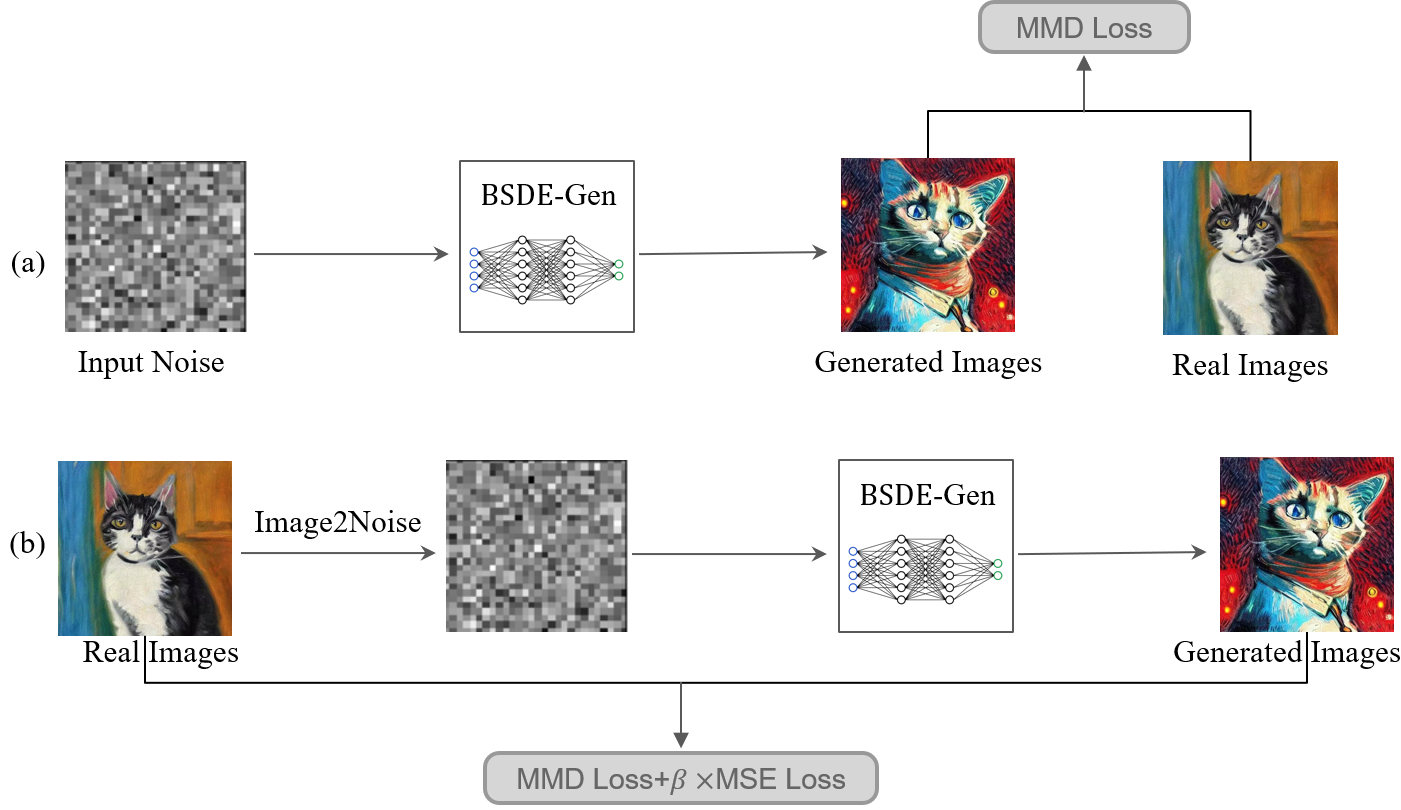}
\par\end{centering}
\caption{Training Strategies for BSDE-Gen}
\label{training_strategy}
\end{figure}
\end{center}
\vspace*{\fill}

\section{Experiments\label{sec:Experiments}}

In this section, we present the experimental results that demonstrate
the effectiveness of our BSDE-based generative models. The models
were trained on two classical image datasets, namely MNIST and FashionMNIST,
each consisting of 60,000 grayscale images of 28$\times$28 pixels.
The MNIST dataset includes images of handwritten digits, while the
FashionMNIST dataset contains images of fashion categories.

In our experiments, the forward state process $X_{t}$ is modeled
as a stationary Ornstein-Uhlenbeck (OU) process starting from $d_{X}=32$
dimensional standard normal distribution $\zeta\sim\mathscr{N}(0,I_{d_{X}})$,
with the drift function $b(t,x)=-x$ and the diffusion function $\sigma(t,x)=\sqrt{2}I_{d_{X}}$.
The generator function $f$ of the backward process $Y_{t}$ is defined
as $f(t,x,y,z)=Ax+By+\kappa|z|$ where $|z|:=(\sum_{j=1}^{d_{W}}|z_{ij}|)_{i=1,2,\cdots,d_{Y}}$,
and $A,B,\kappa$ are given. That is, 
\begin{equation}
\begin{aligned}X_{t} & =\zeta-\int_{0}^{t}X_{s}ds+\sqrt{2}W_{t}.\\
Y_{t} & =\xi+\int_{t}^{T}(AX_{s}+BY_{s}+\kappa|Z_{s}|)ds-\int_{t}^{T}Z_{s}dW_{s}.
\end{aligned}
\end{equation}
Similar BSDEs have been studied in the literature, such as those presented
in \cite{CE2002,CLQX2022}. In our experiments, we set the time step
to $N=200$ and the time horizon to $T=1$.

For the deep neural networks $\mathcal{N}^{\theta_{Y_{0}}}(X_{0})$
and $\mathcal{N}^{\theta_{Z}}(t_{n},X_{t_{n}})$, we utilize a three-hidden-layered
architecture with the GELU activation function and dropout regularization
with probability $p=0.2$. The last mapping function $\phi$ is linear.

We trained our BSDE-based deep generative models using the RMSprop
optimizer with a learning rate of 1e-4, a batch size of 512, and 20,000
epochs with the PyTorch framework using 8 NVIDIA A100 GPUs under both
decoder-only style and encoder-decoder style. We used the multi-scale
kernel for the MMD loss function. We also experimented with several
different hyperparameters, such as the forward process, the generator
function $f$, activation functions, dropout rate, batch size, learning
rate, and optimizer. However, the final performance of the model remained
consistent. This also provides us with the flexibility to employ the
BSDE-based models.

\begin{center}
\begin{figure}[tbph]
\begin{centering}
\includegraphics[width=14cm]{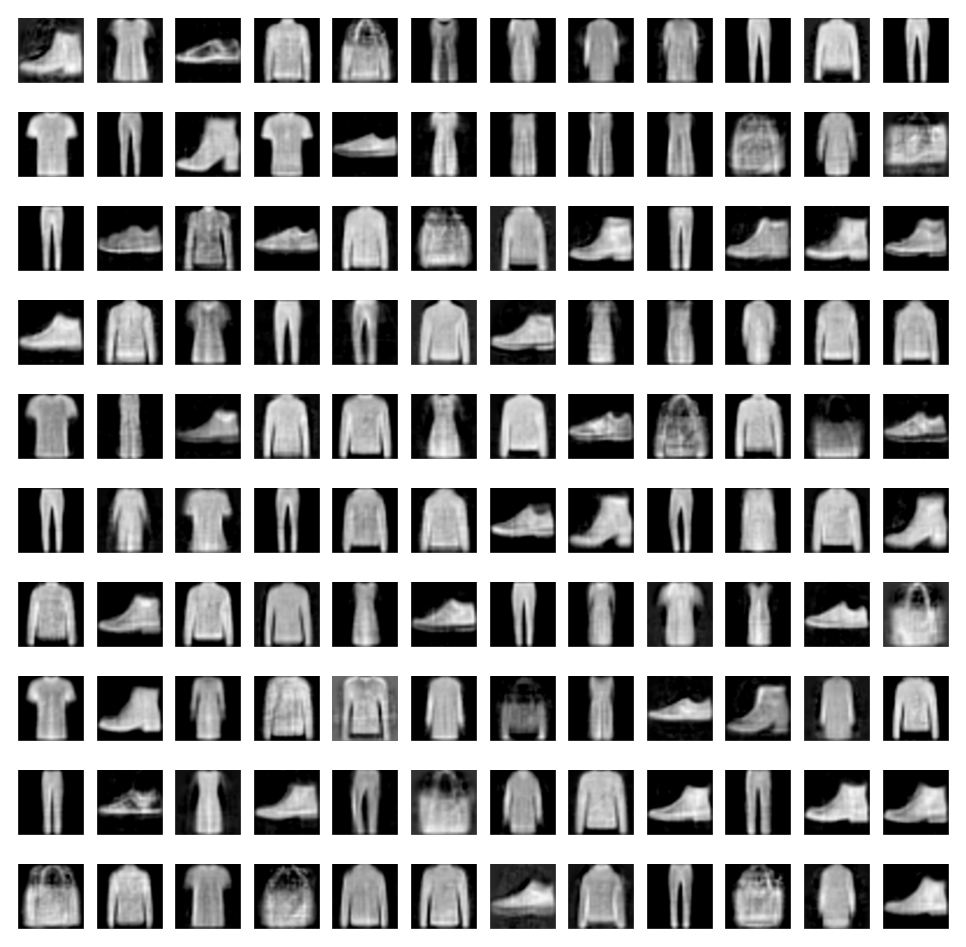}
\par\end{centering}
\caption{Generated images by BSDE-Gen model trained (decoder-only style) on
FashionMNIST using the RMSprop optimizer with a batch size of 512
and a learning rate of 0.0001.}
\label{gen_img_FashionMNIST}
\end{figure}
\end{center}

\begin{center}
\begin{figure}[tbph]
\begin{centering}
\includegraphics[width=14cm]{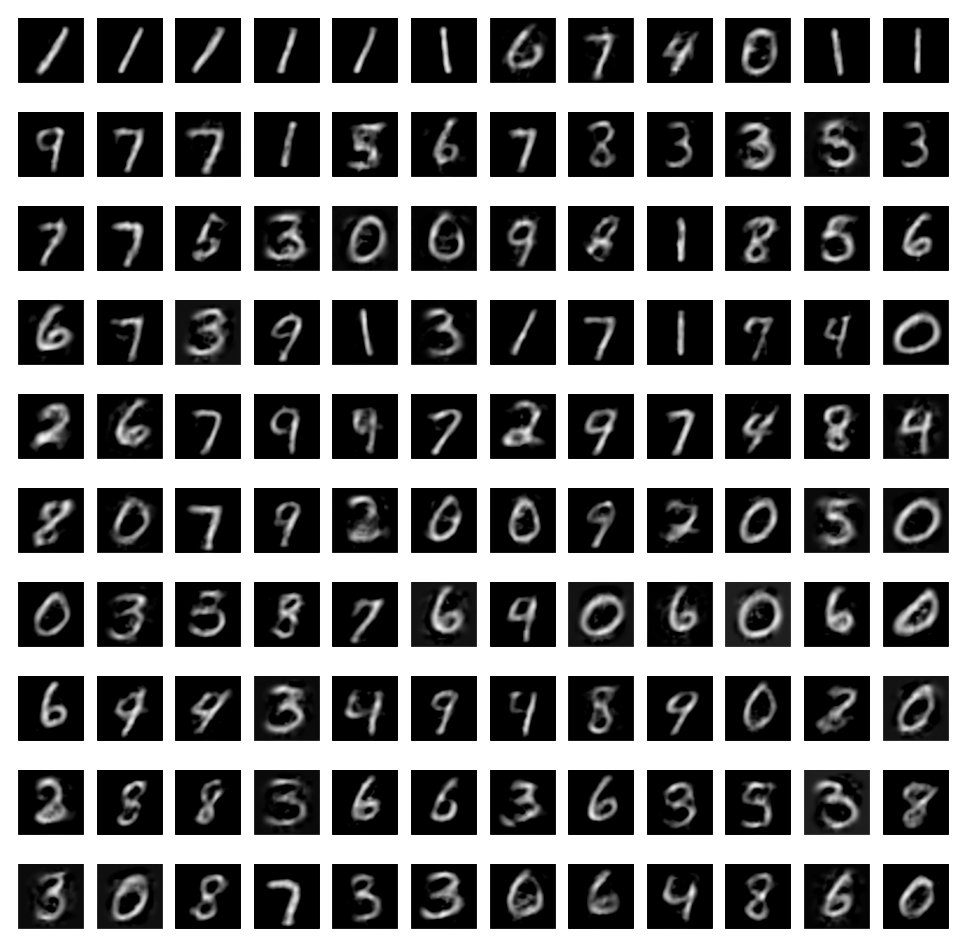}
\par\end{centering}
\caption{Generated images by BSDE-Gen model trained (decoder-only style) on
MNIST using the RMSprop optimizer with a batch size of 512 and a learning
rate of 0.0001.}
\label{gen_img_MNIST}
\end{figure}
\end{center}

\begin{center}
\begin{figure}[!htbp]
\begin{centering}
\begin{minipage}[c]{7.8cm}%
\includegraphics[width=8cm]{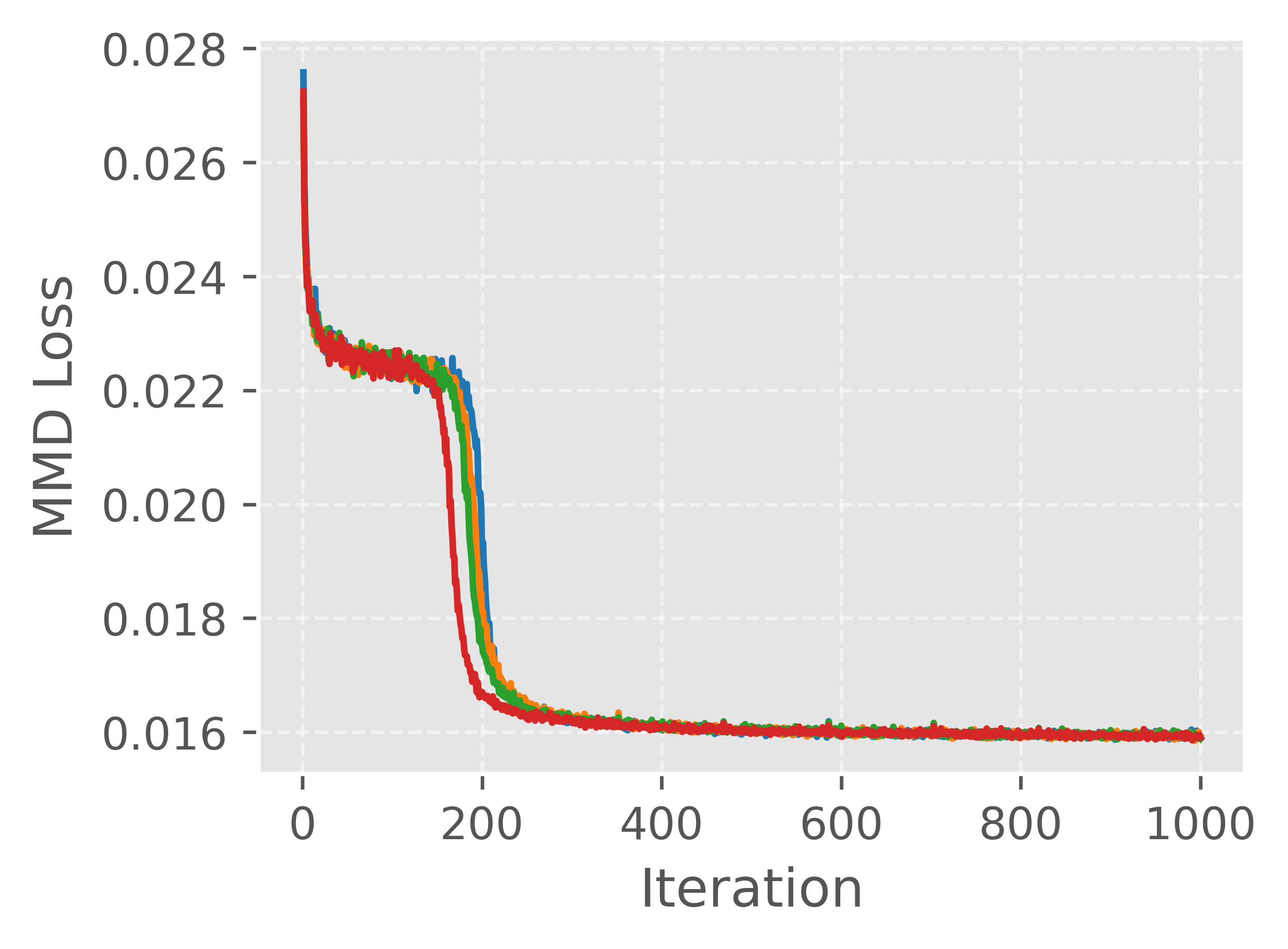}%
\end{minipage}%
\begin{minipage}[c]{7.8cm}%
\includegraphics[width=8cm]{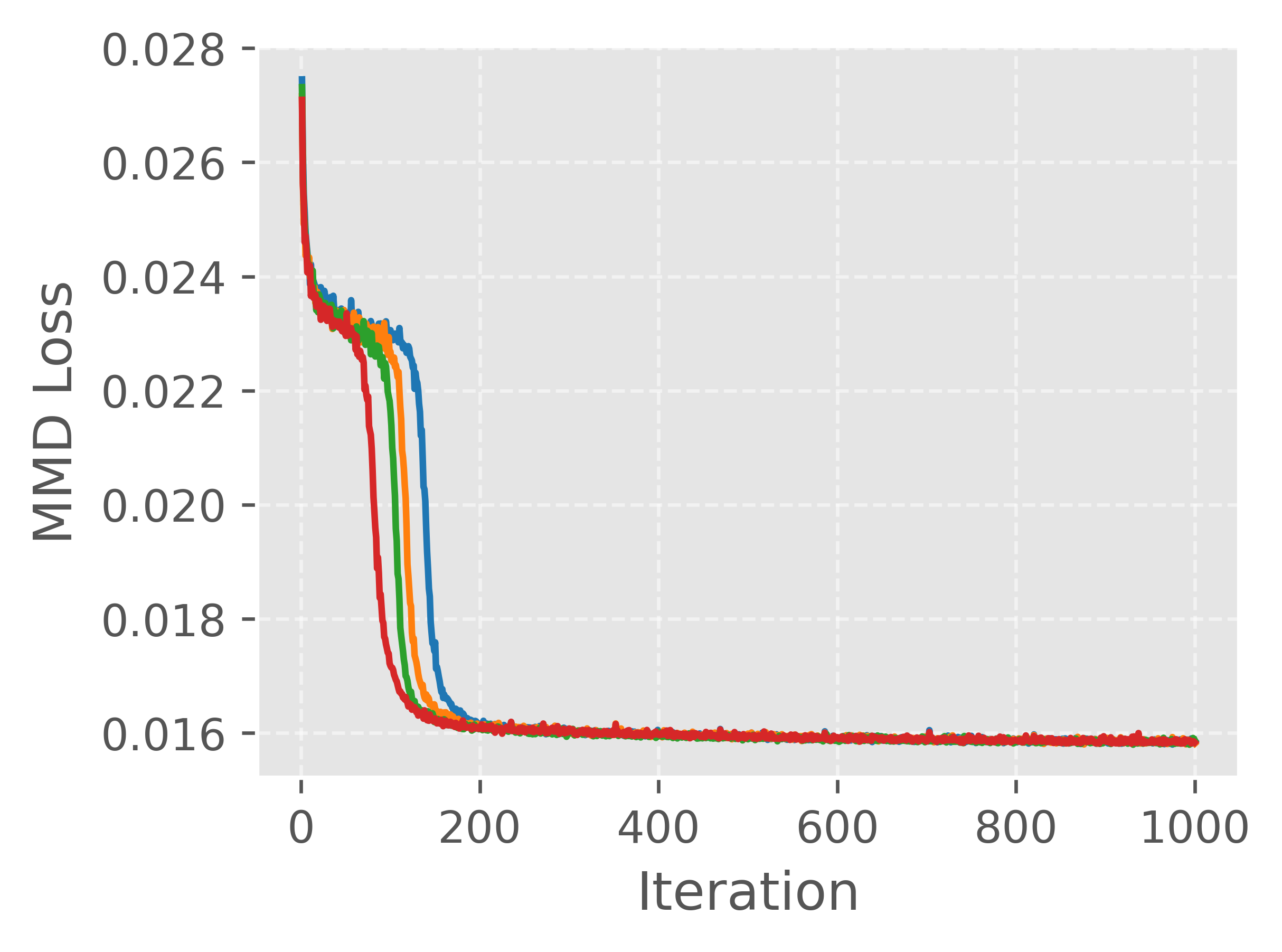}%
\end{minipage}
\par\end{centering}
\begin{centering}
\begin{minipage}[c]{7.8cm}%
\includegraphics[width=8cm]{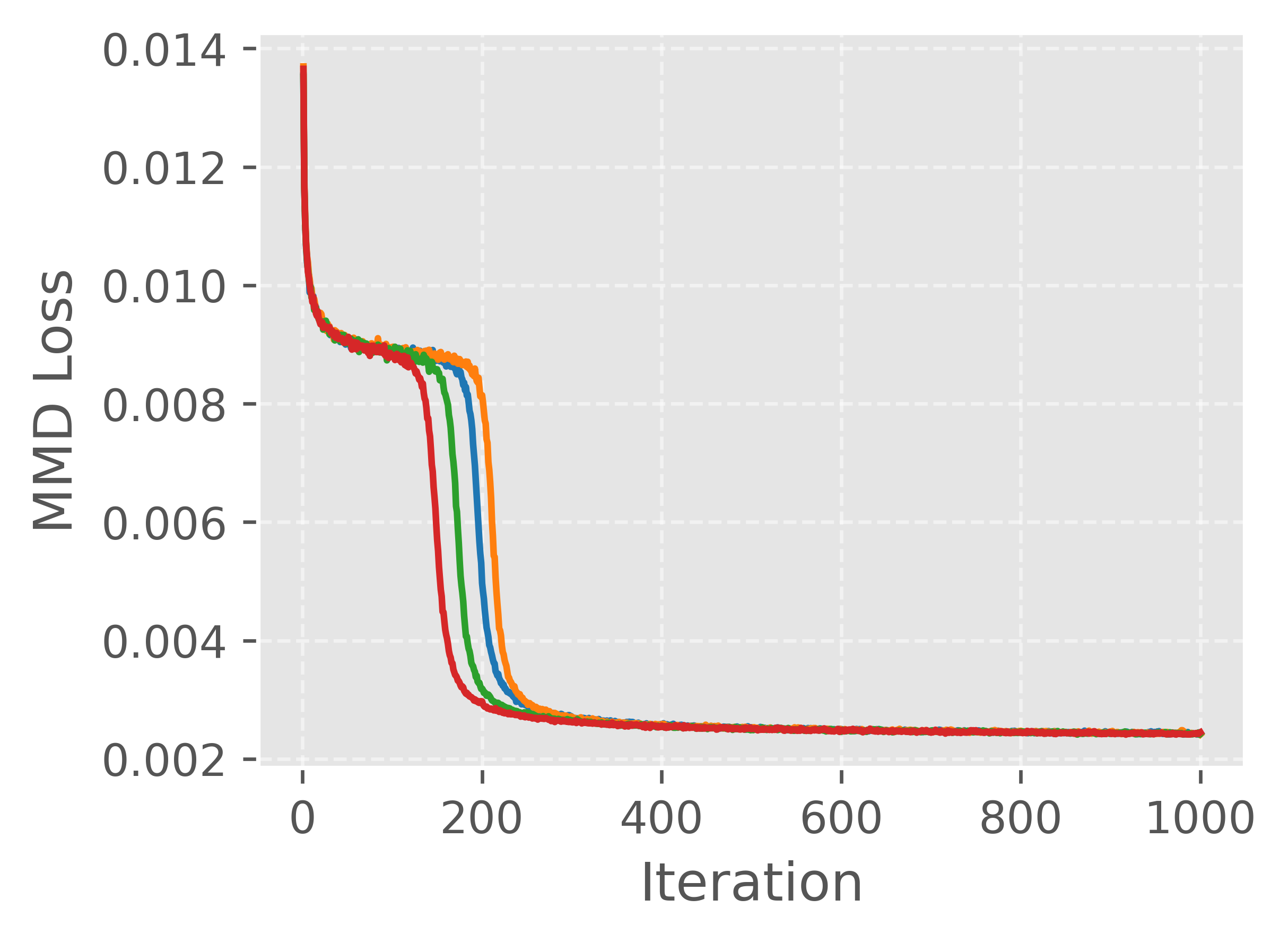}%
\end{minipage}%
\begin{minipage}[c]{7.8cm}%
\includegraphics[width=8cm]{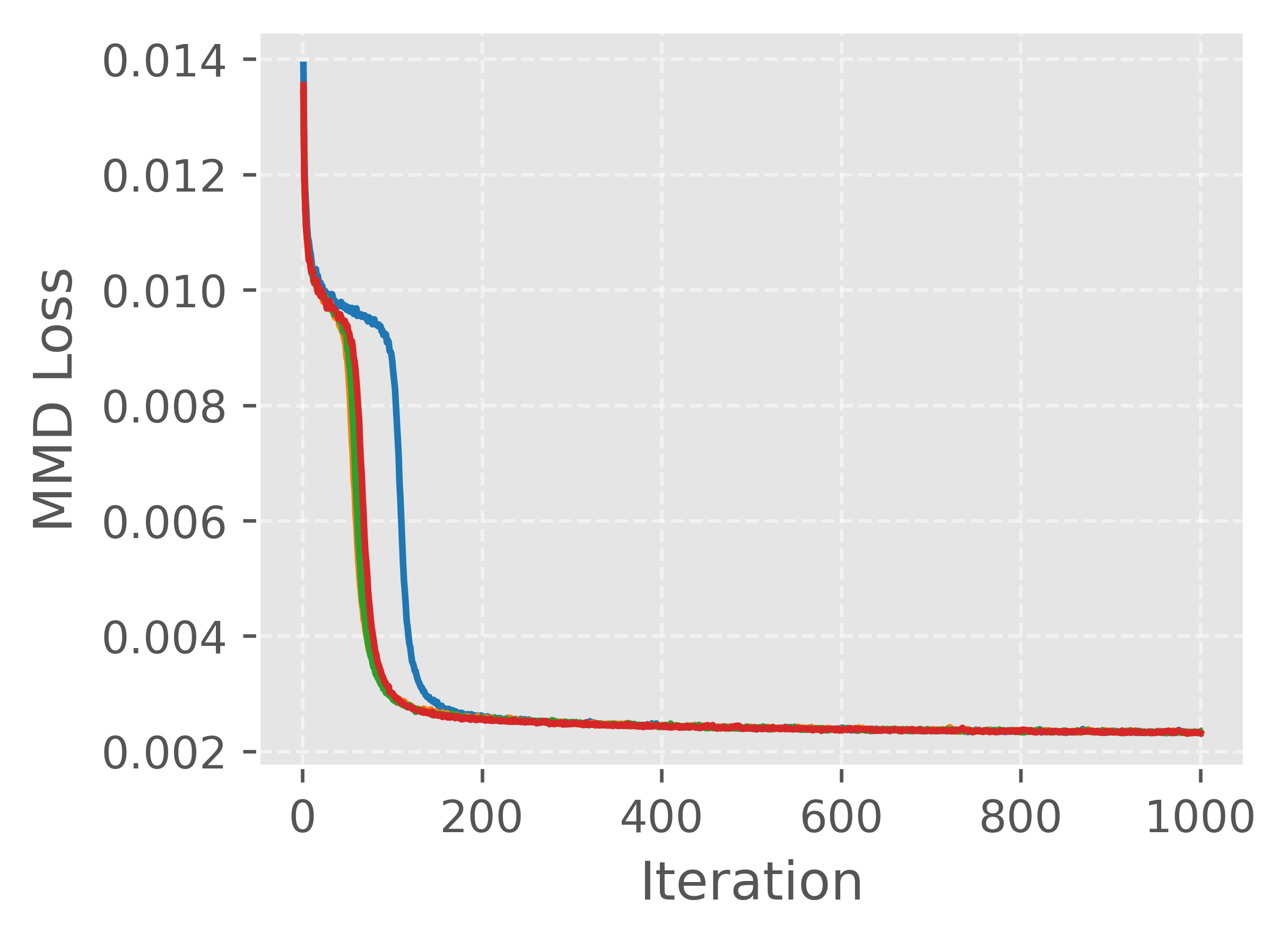}%
\end{minipage}
\par\end{centering}
\caption{Training loss: We ran four training sessions (represented by four different colors) for each panel without fixing the random seed. The RMSprop optimizer with a learning rate of 0.0001
is used. The upper two panels trained (decoder-only style) with a
batch size of 512 and the lower two panels trained with a large batch
size of 3,750. The left panels correspond to FashionMNIST, while the
right panels correspond to MNIST. Only the first 1,000 iterations
are depicted in these figures.}
\label{fig_loss}
\end{figure}
\end{center}

The experiments conducted in this study using the BSDE-Gen models demonstrate
its potential for generative modeling. The generated images are displayed
in Figure \ref{gen_img_FashionMNIST} and \ref{gen_img_MNIST}, and
the models are trained in the decoder-only style for these figures.
Figure \ref{fig_loss} shows the training loss, with the upper two
panels trained with a batch size of 512 and the lower two panels trained
with a large batch size of 3,750. The left panels correspond to FashionMNIST,
while the right panels correspond to MNIST. Only the first 1,000 iterations
are depicted in these figures. As evident from the figures, the BSDE-Gen
models have generated images similar to the training datasets, although
there is still room for improving the visual quality of the generated
images. Future research can explore potential modifications to the
model, including designing the encoder and leveraging a better model
architecture such as U-Net, which we are currently exploring, to enhance
the visual quality. We will discuss these avenues for future research
in more detail in the next section.

\newpage{}

\vspace*{\fill}\vspace*{\fill}

\newpage{}

\vspace*{\fill}\vspace*{\fill}

\newpage{}

\vspace*{\fill}\vspace*{\fill}

\newpage{}

\section{Discussion\label{sec:Discussion}}

In this section, we will discuss the limitations of the BSDE-Gen model
and potential avenues for future research. Although the model presents
a promising approach for generating high-dimensional complex data,
there are some limitations that need to be addressed. Firstly, the
computational complexity of the model can be a bottleneck when working
with large datasets, and optimizing the model for efficiency is an
important future direction. Another limitation is that the BSDE-Gen
model requires a choice of hyperparameters such as the forward process
and the generator function $f$ of the backward process.

Future research on the BSDE-Gen model can focus on developing more
efficient algorithms to reduce computational complexity, and exploring
different encoder architectures that can improve the mapping between the BSDE-Gen 
model input and target image. Additionally, investigating the potential
applications of the BSDE-Gen model in different fields and comparing
its performance with other generative models can further enhance our
understanding of the model's capabilities and limitations. In addition
to our current model architecture, exploring alternative architectures,
such as incorporating the U-Net architecture, could lead to further
improvements in the quality of generated data. Furthermore, the integration
of guided models is an exciting avenue for future research in the
field of BSDE-Gen models, to improve the performance of generated
data by providing additional information or constraints to the generative
model.

\subsection{Diffusion Process Selection}

The BSDE-Gen model is a complex model that requires the selection
of various hyperparameters to achieve optimal performance. Among the
most critical decisions is the choice of the forward process and the
generator function $f$ of the backward process. The forward process
specifies the dynamics of the state, while the generator function of
the backward process specifies how to generate samples from the target
data distribution.

Choosing appropriate hyperparameters is essential to ensure the model's
performance and generate high-quality data. Therefore, a comprehensive analysis and extensive experimentation
are required to determine the optimal settings for the BSDE-Gen model
in a specific application.

\subsection{U-Net Architecture}

The incorporation of the U-Net architecture has demonstrated potential
in enhancing the quality of generated data in several generative models,
as it is capable of capturing both local and global features in images.
Therefore, exploring the implementation of this architecture in the
BSDE-Gen model could lead to further improvements in generated data
quality. Future research could focus on evaluating the performance
of the BSDE-Gen model with the U-Net architecture on larger datasets,
and optimizing its implementation to further enhance its performance.

\subsection{Conditional BSDE-Gen Models}

The integration of guided models is an exciting avenue for future
research in the field of BSDE-Gen models, as it has shown promising
results in improving the performance and quality of generated data
in classical diffusion models. One possible approach for integrating
guided models into our framework would be to use conditional probability
methods to incorporate the guided information into the generation
process. For example, in image generation, the guided information
could be in the form of a segmentation map, a sketch or a descriptive
text, which could be used to condition the generation of the image.
This would involve training the generative model to produce samples
from the conditional distribution of the data given the guidance,
rather than from the unconditional distribution.

\section{Conclusion\label{sec:Conclusion}}

In conclusion, this paper introduces a novel deep generative model
called BSDE-Gen that combines the power of deep neural networks with
the flexibility of BSDEs for generating complex high-dimensional target
data, with a focus on image generation. The incorporation of stochasticity
and uncertainty into the generative modeling process enables the model
to generate data that closely resembles the original dataset. The
paper provides a theoretical foundation for BSDE-Gen, presents its
model architecture, and reports experimental results. The potential
applications of BSDE-Gen in various fields, including computer vision,
biology, and drug discovery, could be significant, as it offers a
new tool for modeling high-dimensional complex systems with uncertain
dynamics and incomplete information. This paper represents a contribution
to the fields of generative modeling, deep BSDE methods, high-dimensional
learning and has the potential to inspire further research in these
areas.
\begin{acknowledgement*}
We thank Zhenyao Sun for many discussions on this subject. Additionally,
we thank ChatGPT for providing assistance in refining the
writing, coding, and debugging throughout the research of this paper.
We also acknowledge the support of Google Colab for providing tools
and resources that facilitated our research.
\end{acknowledgement*}

\end{document}